\documentclass[letterpaper]{article} 
\usepackage{aaai23}  
\usepackage{times}  
\usepackage{helvet}  
\usepackage{courier}  
\usepackage[hyphens]{url}  
\usepackage{graphicx} 
\usepackage{natbib}  
\usepackage{caption} 
\usepackage{algorithm}
\usepackage{algorithmic}

\usepackage{subfigure}
\usepackage{graphicx}
\usepackage{amsmath}
\usepackage{amssymb}
\usepackage{booktabs}
\usepackage{subfig}
\usepackage{bm}
\usepackage{xspace}
\usepackage{multirow}
\usepackage{url}

\makeatletter
\DeclareRobustCommand\onedot{\futurelet\@let@token\@onedot}
\def\@onedot{\ifx\@let@token.\else.\null\fi\xspace}

\def\ie{\emph{i.e}\onedot}

\def\etal{\emph{et al}\onedot}
\makeatother

\usepackage{newfloat}
\usepackage{listings}
\DeclareCaptionStyle{ruled}{labelfont=normalfont,labelsep=colon,strut=off} 
\floatstyle{ruled}
\newfloat{listing}{tb}{lst}{}
\floatname{listing}{Listing}
\title{Improving Training and Inference of Face Recognition Models via Random Temperature Scaling}
\author {
    Lei Shang\textsuperscript{\rm 1,}\equalcontrib,
    Mouxiao Huang\textsuperscript{\rm 2,3,}\equalcontrib,
    Wu Shi\textsuperscript{\rm 2,}\thanks{Corresponding author.},
    Yuchen Liu\textsuperscript{\rm 1},\\
    Yang Liu\textsuperscript{\rm 1},
    Fei Wang\textsuperscript{\rm 1},
    Baigui Sun\textsuperscript{\rm 1,}\thanks{Project lead.},
    Xuansong Xie\textsuperscript{\rm 1},
    Yu Qiao\textsuperscript{\rm 2,4}
}
\affiliations {
    \textsuperscript{\rm 1} Alibaba Group\\
    \textsuperscript{\rm 2} The Guangdong Provincial Key Laboratory of Computer Vision and Virtual Reality Technology, \\ Shenzhen Institute of Advanced Technology, Chinese Academy of Sciences\\
    \textsuperscript{\rm 3} University of Chinese Academy of Sciences\\
    \textsuperscript{\rm 4} Shanghai Artificial Intelligence Laboratory\\
    \{sl172005, yuchen.lyc, ly261666, steven.wf, baigui.sbg, xingtong.xxs\}@alibaba-inc.com\\
    \{mx.huang, wu.shi, yu.qiao\}@siat.ac.cn
}

\usepackage{bibentry}

\begin{document}

\maketitle

\begin{abstract}
Data uncertainty is commonly observed in the images for face recognition (FR). However, deep learning algorithms often make predictions with high confidence even for uncertain or irrelevant inputs. Intuitively, FR algorithms can benefit from both the estimation of uncertainty and the detection of out-of-distribution (OOD) samples. Taking a probabilistic view of the current classification model, the temperature scalar is exactly the scale of uncertainty noise implicitly added in the softmax function. Meanwhile, the uncertainty of images in a dataset should follow a prior distribution. Based on the observation, a unified framework for uncertainty modeling and FR, Random Temperature Scaling (RTS), is proposed to learn a reliable FR algorithm. The benefits of RTS are two-fold. (1) In the training phase, it can adjust the learning strength of clean and noisy samples for stability and accuracy. (2) In the test phase, it can provide a score of confidence to detect uncertain, low-quality and even OOD samples, without training on extra labels. Extensive experiments on FR benchmarks demonstrate that the magnitude of variance in RTS, which serves as an OOD detection metric, is closely related to the uncertainty of the input image. RTS can achieve top performance on both the FR and OOD detection tasks. Moreover, the model trained with RTS can perform robustly on datasets with noise. The proposed module is light-weight and only adds negligible computation cost to the model.
\end{abstract}

\begin{figure}[t]
   \centering
   \includegraphics[width=0.95\linewidth]{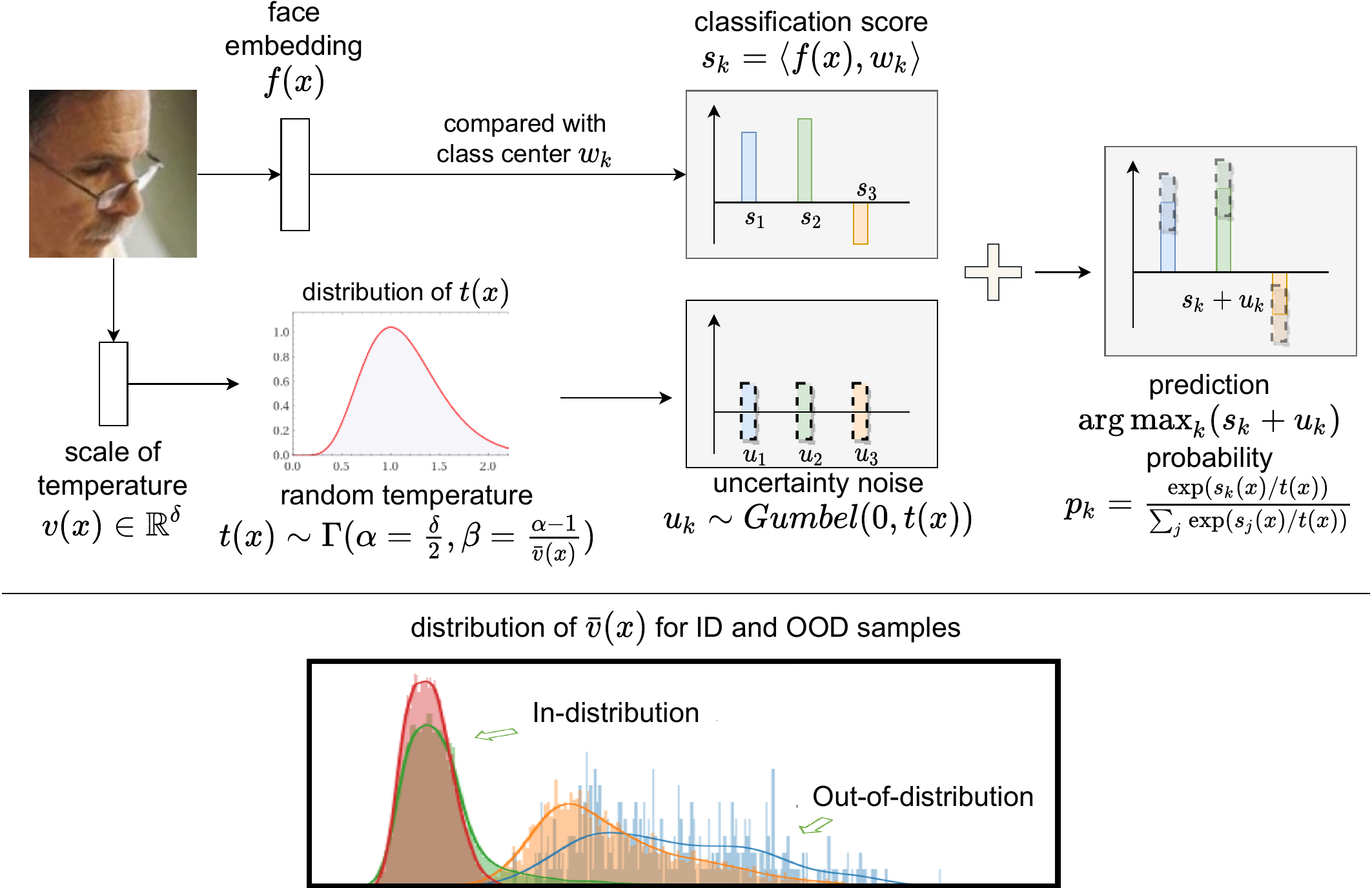}
   \caption{Random Temperature Scaling models the data uncertainty via the scale of temperature (top). We find that the learned scale, $v(\bm{x})$, is a good metric for out-of-distribution detection (bottom).}
\label{fig:intro}
\end{figure}

\section{Introduction}
\label{sec:intro}
Recently, deep neural networks have achieved great success in face recognition. State-of-the-art face recognition methods generally map each face image to an identity-related representation in the latent space~\cite{wen2016discriminative,deng2019arcface,wang2018cosface,schroff2015facenet,liu2017sphereface}. 
Ideally, the representations of the same person are embedded in a compact cluster. However, when the face image contains uncertainty and noise, the learned representation may be unreliable and tend to cause mistakes.

It is well known that deep learning models can achieve high accuracy in many classification tasks given a relatively clean and closed dataset. 
However, in the real world, the data are captured in more unconstrained scenarios~\cite{shi2016deep,wang2020suppressing}. Data uncertainty commonly present in the images for face recognition. For example, in surveillance videos~\cite{kalka2018ijb,maze2018iarpa,klare2015pushing}, the image can be corrupted by motion blur, focus blur, and lighting conditions. When key facial landmarks are occluded due to pose, hair, or accessories, it is not confident to extract accurate representations for matching. In the step of face detection, partial face or non-face images can be proposed for recognition. The results for these data can be unpredictable. 
It is essential to estimate the uncertainty of data and detect OOD samples for safety-critical algorithms like face recognition.

Probabilistic face embeddings (PFE)~\cite{shi2019probabilistic} is the first to consider data uncertainty in face recognition. Instead of a fixed representation, the face is embedded as a Gaussian distribution whose mean is the original representation and variance is estimated by another model. A new similarity metric, mutual likelihood score (MLS), is proposed to measure the matching likelihood between two images. 
The main limit of PFE is that it can not learn the representation at the same time. 
Data uncertainty learning (DUL)~\cite{chang2020data} also models the face representation as a Gaussian distribution. It uses the re-parameterization trick to simultaneously learn the feature (mean) and uncertainty (variance). The learned features can be directly used for conventional similarity metrics with better intra-class and inter-class properties. The learned uncertainty affects the model by adaptively reducing the adverse effects of noisy images.

Current state-of-the-art OOD methods are generally designed for classification tasks. ODIN~\cite{liang2018enhancing} provides two strategies, temperature scaling and input preprocessing, for OOD detection, based on a trained neural network classifier. It finds that the maximum class probability (called \emph{softmax score}) is an effective score for detecting OOD data. Despite its effectiveness, ODIN requires OOD data to tune hyperparameters. This requirement could prevent the tuned hyperparameter to generalize to other datasets~\cite{shafaei2019less}. 
GODIN~\cite{hsu2020generalized} extends the setting of ODIN, and proposes two strategies, data-dependent temperature scaling and auto-tuned input preprocessing, for learning without OOD data.
Techapanurak \etal~\cite{techapanurak2019hyperparameter} propose a hyperparameter-free method based on softmax of scaled cosine similarity. The proposed method shows at least comparable performance to the state-of-the-art methods on the conventional test, and outperforms the existing methods on the recently proposed test~\cite{shafaei2019less}.
Although these methods are not designed for face recognition models, it is worth conducting experiments to evaluate the performance on OOD detection when learning face representations at the same time. 

We propose an effective method, Random Temperature Scaling (RTS), to learn a reliable face recognition model.
First we take a probabilistic view of the classification method using softmax with temperature scaling, and find that the temperature scalar is exactly the scale of uncertainty noise implicitly used in softmax. The larger the temperature is, the larger the variance of uncertainty becomes. 
Instead of a fixed scalar, RTS models the stochastic distribution of temperature which depends on the input data. RTS acts like a Bayesian network~\cite{gal2016dropout,theobald2021bayesian} and has a regularization effect on the confidence.
Both the face representation and the temperature can be learned simultaneously. Besides, RTS adds only negligible computation cost to the original face recognition model.
Our contributions include:
(1) We reveal the connection between temperature scaling and uncertainty in classification and propose Random Temperature Scaling (RTS) to learn reliable face recognition models.
(2) In the training phase, RTS can adjust the learning strength of clean and noisy samples for stability and accuracy. 
(3) In the test phase, RTS can provide an uncertainty score to detect uncertain, low-quality and even OOD samples, without training on extra labels.

\section{Related Work}
\label{sec:related}

\paragraph{Uncertainty in face recognition.}
Uncertainty in deep learning models is getting more attention these years. How uncertainty affects the performance of deep neural networks is investigated in many works~\cite{gal2016dropout,blundell2015weight,kendall2017uncertainties}.
The pioneering work~\cite{kendall2017uncertainties} introduces two types of uncertainty for deep neural networks via Bayesian modeling. Data uncertainty captures noise in the data distribution, and model uncertainty accounts for uncertainty in the model parameters. Model uncertainty can be explained away given enough data. 
In the real world, data uncertainty commonly presents in the images for face recognition. Modeling data uncertainty is essential for reliable deep learning models. 
PFE~\cite{shi2019probabilistic} is the first work to consider data uncertainty in face recognition task. In PFE, each face is embedded as a Gaussian distribution whose mean is the original representation and variance is estimated by another model. The main limit of PFE is that it relies on the pre-trained face representations. 
DUL~\cite{chang2020data} is the first work to train the face representation and uncertainty at the same time. 

\paragraph{Out-of-distribution detection.}
It is known that deep learning algorithms often make high confidence predictions even for unrecognizable or irrelevant objects~\cite{moosavi2017universal,nguyen2015deep}. Besides capturing such uncertainty in classifiers, detecting the OOD sample is a more direct approach to that problem. 
ODIN~\cite{liang2018enhancing} finds that the maximum class probability (called \emph{softmax score}) is an effective score for detecting OOD data, and provides two strategies, temperature scaling and input pre-processing, for OOD detection. Despite its effectiveness, ODIN requires OOD data to tune hyperparameters.
GODIN~\cite{hsu2020generalized} extends the setting of ODIN, and proposes two corresponding strategies for learning without OOD data.
Techapanurak \etal~\cite{techapanurak2019hyperparameter} propose a hyperparameter-free method based on softmax of scaled cosine similarity, and experiments show its competitive performance in many OOD tests.

\paragraph{Temperature scaling.} 
Temperature scaling is a widely used technique for confidence calibration~\cite{guo2017calibration,neumann18relaxed}, knowledge distillation~\cite{hinton2015distilling}, and adaptive classification objectives~\cite{zhang2019adacos}.
In knowledge distillation~\cite{hinton2015distilling}, the temperature scalar is a global value tuned for each task and each model to soften the prediction.
In~\cite{zhang2019adacos}, the temperature is dynamically adjusted to learn a better representation.
In confidence calibration~\cite{neumann18relaxed}, the temperature is learned depending on the input data.
We find the temperature scalar is connected to the scale of uncertainty in the classifier.

\begin{figure}[t]
\centering
   \includegraphics[width=0.9\linewidth]{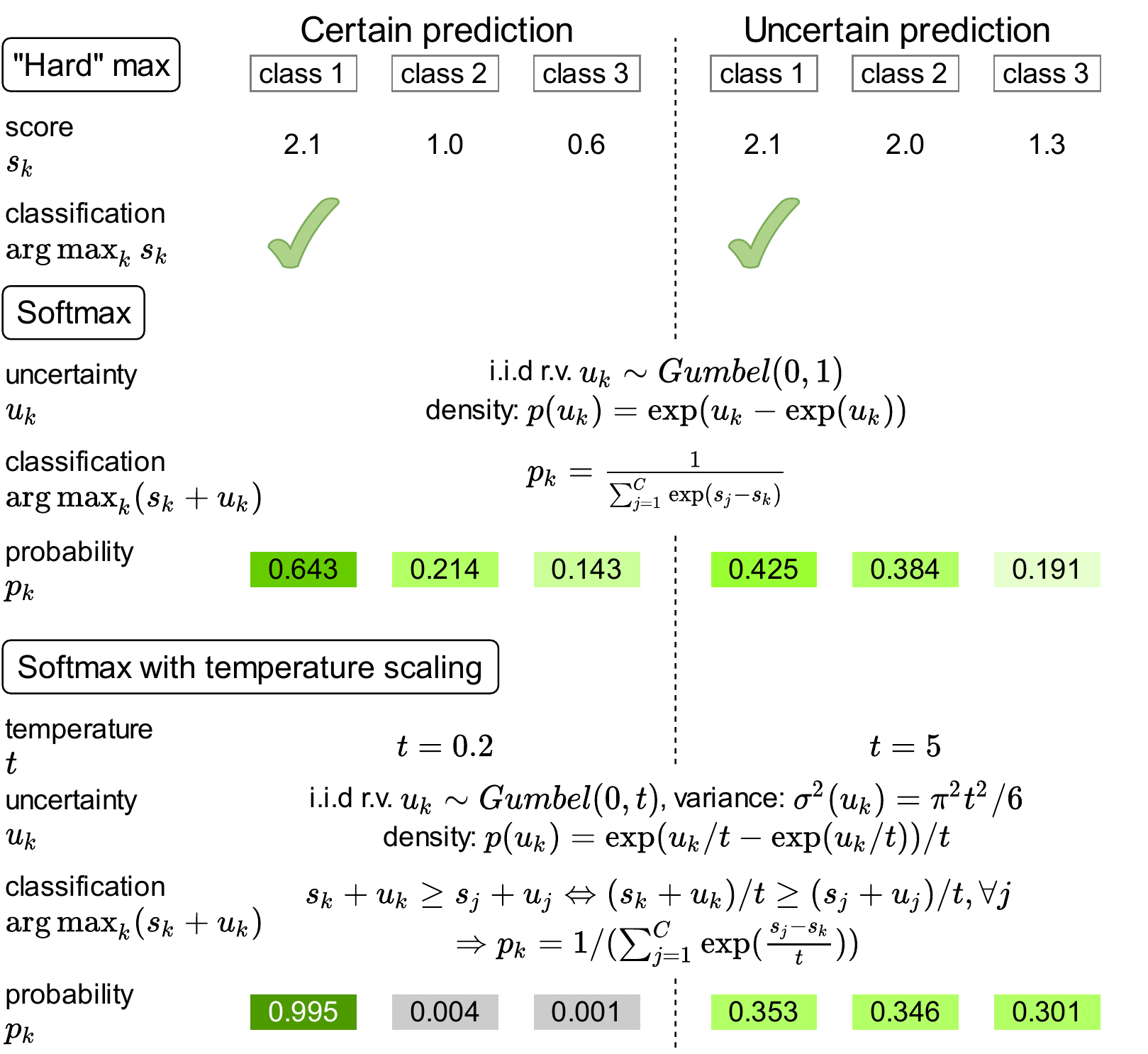}
   \caption{A probabilistic view of softmax and temperature scaling. The standard softmax function models the uncertainty of prediction by implicit Gumbel random variables. In temperature scaling, the temperature corresponds to the scale of uncertainty noise added to the classification score.}
\label{fig:temp-scaling}
\end{figure}

\section{Methodology}
\label{sec:method}

\subsection{Preliminaries}
\label{ssec:preliminary}
A face recognition algorithm generally maps an image $\bm{x} \in \mathbb{R}^{h\times w\times 3}$ to an identity-related representation $\bm{y} = f(\bm{x}) \in \mathbb{R}^{d_y}$. 
The predicted probability is defined by
\begin{equation}
  p_k(\bm{x}) = \frac{\exp (s_k(\bm{x}))}{\sum_j \exp(s_j(\bm{x}))},~~~~s_j(\bm{x}) = \langle f(\bm{x}), \bm{w}_j\rangle,
  \label{eq:softmax}
\end{equation}
where $\bm{w}_j$ is called the class center for identity $j \in \{1, \dots, C\}$ and $C$ is the number of different classes.
Given a set of images $\{\bm{x}_1,\dots, \bm{x}_N\}$ and their labels $\{c(\bm{x}_1), \dots, c(\bm{x}_N)\}$, the classifier is trained by minimizing the Cross Entropy loss:
\begin{equation}
  \mathcal{L}_{CE} = -\frac{1}{N} \sum_{i = 1}^{N} \log \left(p_{c(\bm{x}_i)}(\bm{x}_i)\right).
  \label{eq:cross_entropy}
\end{equation}

To improve the discriminative ability of the learned features, various margin-enhanced losses~\cite{deng2019arcface,wang2018cosface} are proposed. 
For example, the ArcFace function $s_k = ArcFace(\bm{y}, c, \{\bm{w}_j\}, k; \gamma, \theta)$ is defined by: 
\begin{equation}
  s_k =
  \begin{cases}
    \gamma \cdot \cos( \cos^{-1}( \langle \bm{y}, \bm{w}_k \rangle) + m), & k = c\\
    \gamma \cdot \langle \bm{y}, \bm{w}_k \rangle, & k \neq c
  \end{cases},
  \label{eq:arcface}
\end{equation}
where $c$ is the corresponding label of $\bm{y}$, $\gamma$ is the scaling factor and $m$ is the additive arc margin. 
Substituting Eqn.~\ref{eq:arcface} into Eqn.~\ref{eq:softmax}, we can derive the predicted probability by ArcFace.
%

\subsection{Random Temperature Scaling}
\label{ssec:rts}

In the following, we elaborate on the principles and details of the proposed method, Random Temperature Scaling (RTS), for face recognition with uncertainty.

\paragraph{A probabilistic view of softmax and temperature scaling.}
As shown in Fig.~\ref{fig:temp-scaling}, a typical classification model predicts the score, $s_k$, for each possible class. To make a decision, the model chooses the class with the maximum score, which can be regarded as a ``Hard'' max method (top).
However, this method is deterministic and cannot indicates the confidence of results. Intuitively, the prediction made from scores (2.1, 1.0, 0.6) should be more confident than that from (2.1, 2.0, 1.3), though the two cases result in the same prediction. 
To consider the confidence, a noise, $u_k$, can be added to the score, and the model predicts from the score $s_k + u_k$ with uncertainty (middle). Assuming that $u_k$'s are i.i.d. Gumbel random variables, the probability of class $k$ is given by:
\begin{equation}
  \Pr(s_k+u_k \geq s_j+u_j, \forall j) = \frac{1}{\sum_{j = 1}^{C} \exp(s_j - s_k)}.
  \label{eq:prob-softmax}
\end{equation}
The scale of uncertainty noise can vary by a factor $t$, \ie, $u_k\sim Gumbel(0, t)$, to reflect the different uncertainties of inputs. Through the change of variables, we can obtain
\begin{equation}
  \Pr(\frac{s_k+u_k}{t} \geq \frac{s_j+u_j}{t}, \forall j) = \frac{1}{\sum_{j = 1}^{C} \exp(\frac{s_j - s_k}{t})}.
  \label{eq:prob-softmax-ts}
\end{equation}
Eqn.~\ref{eq:prob-softmax-ts} is exactly the softmax function with temperature scaling. The temperature corresponds to the scale of uncertainty noise, $u_k$, added to the classification score (Fig.~\ref{fig:temp-scaling} bottom). The variance of $u_k$ is $\pi^2 t^2 / 6$. The larger the temperature is, the larger the variance of uncertainty becomes. 

\begin{algorithm}[tb]
\caption{Random Temperature Scaling (RTS)}
\label{alg:rts}
\textbf{Input}: Image $\bm{x}$, label $c$, dof $\delta$, scale $\gamma$, margin $m$ \\
\textbf{Parameter}: $\theta_f, \theta_g, \{\bm{w}_j\}$ \\
\textbf{Output}: Scores $\bm{s}$
\begin{algorithmic}[1] 
\STATE face repr. $\bm{y} = f(\bm{x}; \theta_f)$\
\STATE log scale $z = g(\bm{x}; \theta_g) \in \mathbb{R^\delta}$\
\STATE scale $v = \exp (z) \in \mathbb{R^\delta}$\
\STATE $\epsilon_i \sim \mathcal{N}(0, 1), \forall i = 1, \dots, \delta$\
\STATE $t = \frac{1}{\delta - 2} \cdot \sum_i v_i\cdot \epsilon_i^2$\
\STATE $s_k^\prime = ArcFace(\bm{y}, c, \{\bm{w}_j\}, k; \gamma, \theta), \forall k = 1, \dots, C$\
\STATE $\bm{s} = [s_1^\prime / t, \dots, s_C^\prime / t]$\
\end{algorithmic}
\end{algorithm}

\begin{figure}[t]
\centering
   \includegraphics[width=0.47\linewidth]{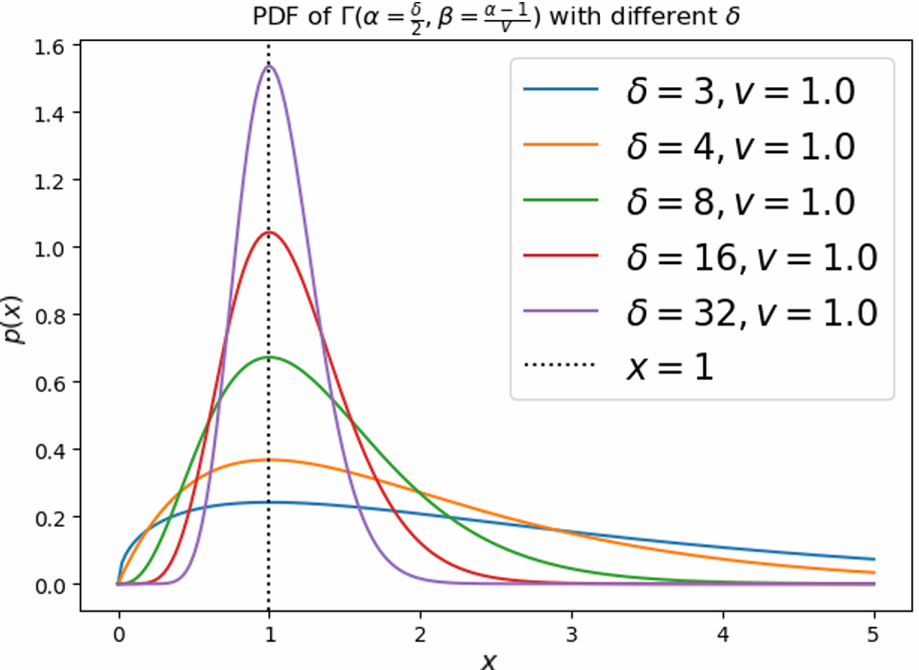}
   \includegraphics[width=0.47\linewidth]{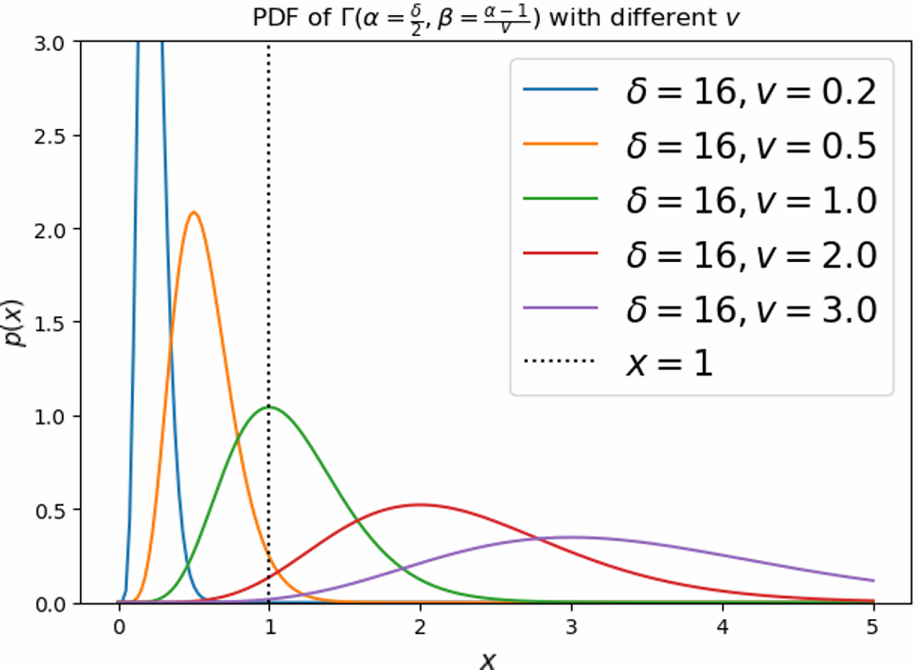}
   \caption{The distributions of $t$ with different $\delta$ and $v$.}
\label{fig:temp-dist}
\end{figure}

\paragraph{A stochastic distribution of the temperature.} 
We assume that the temperature in a dataset follows a prior distribution, we propose the Random Temperature Scaling (RTS) method to model the stochastic distribution of temperature. In RTS, the temperature, $t(x)$, is a random variable whose scale depends on the input data.
The algorithm is described in Alg.~\ref{alg:rts}. Specifically, $f$ is the head for face representation and $g$ is the head for log scale of the temperature scalar. 
The temperature is a sum of $\delta$ independent Gamma variables: $r_i = v_i \cdot \epsilon_i^2 \sim \Gamma(\frac{1}{2}, \frac{1}{2v_i})$. For illustration, we let $v_i = v, \forall i$. We divide the sum by $(\delta - 2)$ to let the mode (most frequent value) of $t$ to be equal to $v$. Thus, $t$ follows $\Gamma(\alpha = \frac{\delta}{2}, \beta = \frac{\alpha - 1}{v})$. The effects of $\delta$ and $v$ are shown in Fig.~\ref{fig:temp-dist}. 
The degree of freedom, $\delta$, is a hyperparameter and the scale, $v(x)$, is a learned function of the input image.

\begin{figure}[t]
\centering
  \includegraphics[width=0.95\linewidth]{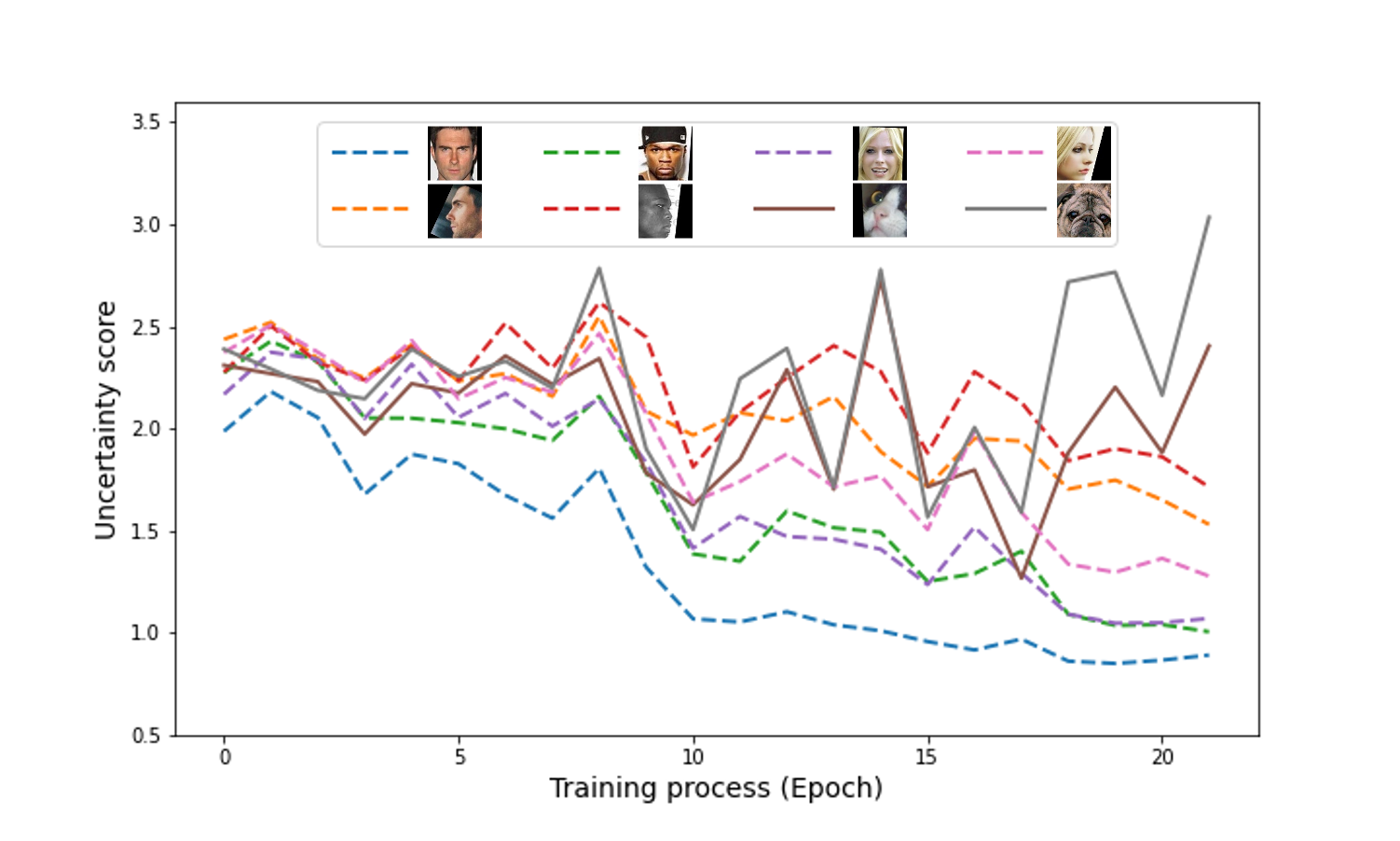}
  \caption{Variation of uncertainty score in RTS during the training process. Training data: DeepGlint.}
\label{fig:variation_uncertainty_curves}
\end{figure}

\paragraph{Difference between Relaxed Softmax and RTS.}
Relaxed Softmax~\cite{techapanurak2019hyperparameter,neumann18relaxed} learns to predict a fixed uncertainty level for each image, while RTS models the uncertainty level by a stochastic distribution.
Relaxed Softmax tends to increase the uncertainty level at early stage and doesn't turn back the trend to result in an overly smoothed prediction. 
RTS has a regularization effect on the confidence. The dynamic of temperature during training is depicted in Fig.~\ref{fig:variation_uncertainty_curves}.
The distribution gradually becomes close to its prior. This encourages the classifier to make a correct prediction, instead of simply tuning down its confidence.

\paragraph{Difference between DUL and RTS.}
DUL~\cite{chang2020data} adds explicit uncertainty noise to the feature embedding, while RTS employs the implicit noise in the softmax function to model the uncertainty. 
For DUL, adding randomness into the representation leads to ambiguity of learning objective~\cite{tolstikhin2018wasserstein}, and sampling from the high dimensional distribution is not efficient for training~\cite{kingma2014auto}. RTS mildly changes the confidence, instead of changing the predicted result.

\subsection{Training Objectives}
\label{ssec:objective}

We train the classifier using softmax with RTS.
In addition to the face feature $f(\bm{x})$, the classifier also predict a log scale, $g(\bm{x}) = \log v(\bm{x}) \in \mathbb{R}^{\delta}$, of the underlying Gamma variable. In our implementation, $f$ and $g$ share the same backbone neural network and use different prediction heads. 
The scaled logits, $\bm{s}$, are given in  Alg.~\ref{alg:rts}.
The Cross Entropy loss, $\mathcal{L}_{CE}$, can be obtained by substituting $\bm{s}$ into Eqn.~\ref{eq:cross_entropy}.
%
Besides, we add a constraint that encourages each independent Gamma variable, $r_i$, to follow the prior distribution $\Gamma(\frac{1}{2}, \frac{1}{2})$:
\begin{equation}
  \begin{aligned}
    \mathcal{L}_{KL} = & \frac{1}{\delta} \sum_{i=1}^{\delta} D_{KL}\left(\Gamma(\frac{1}{2}, \frac{1}{2v_i}), \Gamma(\frac{1}{2}, \frac{1}{2})\right) \\
    = &\frac{1}{\delta} \sum_{i=1}^{\delta} \frac{1}{2}(v_i - \log v_i - 1)
  \end{aligned}
  \label{eq:loss_kl}
\end{equation}
Finally, the overall loss is given by
\begin{equation}
  \mathcal{L} = \mathcal{L}_{CE} + \lambda\cdot \mathcal{L}_{KL},
\end{equation}
where $\lambda$ is a hyperparameter to balance the two losses. 

\section{Experiments}
\label{sec:experiments}

\subsection{Implementation Details}
\label{ssec:details}

\paragraph{Datasets.}
We use CASIA-WebFace (10k IDs / 0.5M Images)~\cite{yi2014learning} and DeepGlint (181k IDs / 6.75M Images)~\cite{deepglint} as the training data for our experiments. For face recognition task, seven benchmarks including LFW~\cite{huang2008labeled}, CFP~\cite{sengupta2016frontal} (frontal-frontal and frontal-profile protocol), AgeDB~\cite{moschoglou2017agedb}, CALFW~\cite{zheng2017cross}, CPLFW~\cite{zheng2018cross}, and VGG2~\cite{cao2018vggface2} are used to evaluate the performance of the baseline and our methods. For OOD detection task, we build an OOD test set consisting of human face, cat and dog images. For in-distribution data, we use LFW~\cite{huang2008labeled} and random select 15339 images from RAFDB~\cite{li2017reliable}. All the face images are not seen in the training phase. For out-of-distribution data, we collect 349 dog images and 3671 cat images. It is worth noting that all the OOD images are misclassified by a face detector as face images. Examples are shown in Fig.~\ref{fig:variation_uncertainty_curves}.

\paragraph{Training details.}
We use ResNet~\cite{he2016deep} backbone with SE~\cite{hu2018squeeze} blocks as the baseline model (Backbone: ResNet64). The dimension of face embedding is 512. All the models use the same backbone. Baseline, GODIN~\cite{hsu2020generalized}, Relaxed Softmax~\cite{techapanurak2019hyperparameter,neumann18relaxed}, DUL~\cite{chang2020data}, and RTS use ArcFace~\cite{deng2019arcface} as the prediction head while MagFace~\cite{meng2021magface} and AdaFace~\cite{kim2022adaface} uses their own prediction heads. Our model (RTS) has extra output dimensions which corresponds to the log-variance $g(\bm{x})$ of the underlying Gamma variable. According to the characteristics of methods, we divide all models into two categories: non-uncertainty models (Baseline, MagFace and AdaFace) and uncertainty models (GODIN, Relaxed Softmax, DUL and RTS).
We follow ArcFace~\cite{deng2019arcface} and DUL~\cite{chang2020data} to set experimental parameters. On CASIA-WebFace, the initial learning rate is 0.1, and is decreased to 0.01 and 0.001 after epoch 20 and 28. The training process is finished at epoch 32. On DeepGlint, models are trained for 22 epochs. The learning rate starts from 0.1 and is divided by 10 at epoch 10 and epoch 18. We use the SGD optimizer with a momentum of 0.9, weight decay of 0.0005 and batch size of 512. Besides, during the experiments, we found that the convergence of uncertainty models and variation of distributions of dataset during the training process are sensitive to the margin in ArcFace especially for Relaxed Softmax. Small margin at the beginning makes models easier to converge. Thus the margin changes linearly in the former half of the training process and remains unchanged in the latter half of the training process. And it is worth noting that this training trick has no obvious impact on the final performance (both verification benchmarks and out-of-distribution metrics). In order to more rigorously eliminate the impact of this training trick, in our comparative experiments, all models with ArcFace prediction head uses dynamic margin without particular instructions.

\begin{figure*}[t]
\centering
  \includegraphics[width=0.16\linewidth]{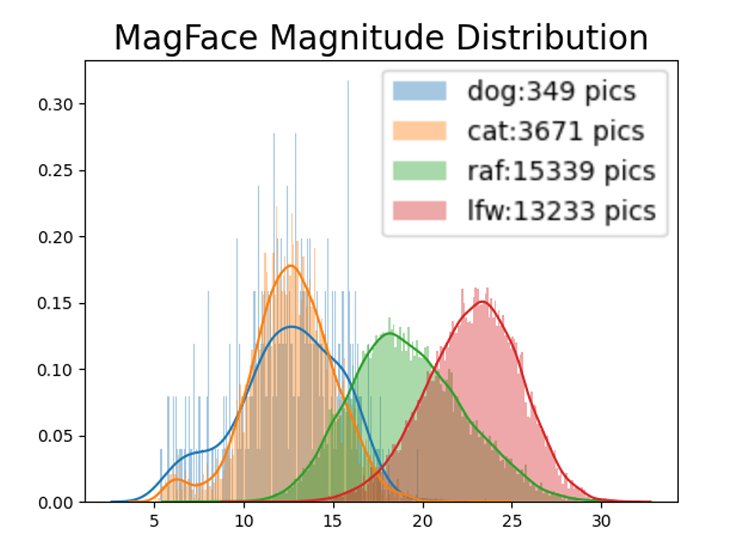}
  \includegraphics[width=0.16\linewidth]{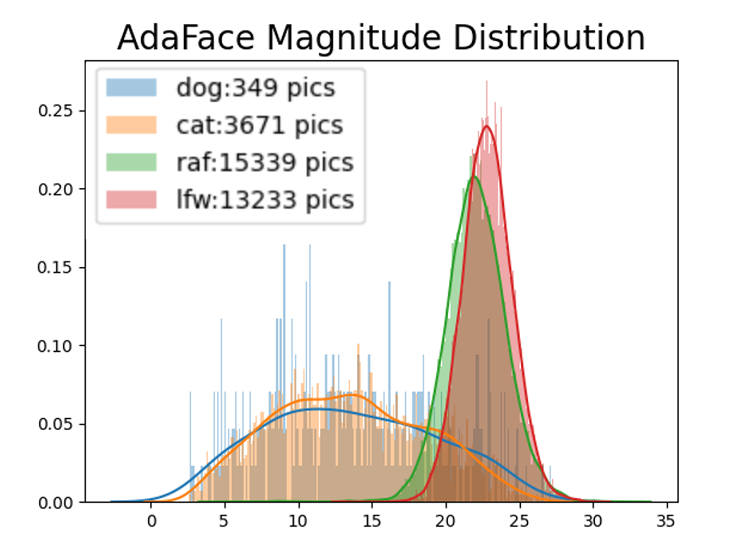}
  \includegraphics[width=0.16\linewidth]{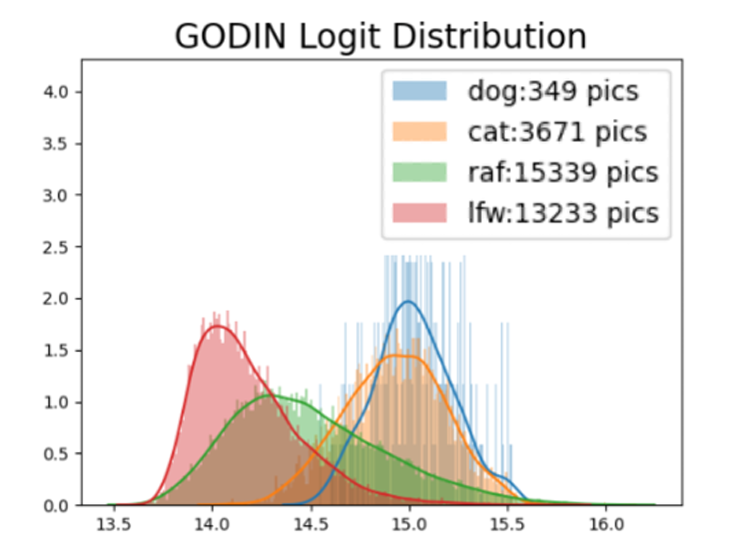}
  \includegraphics[width=0.16\linewidth]{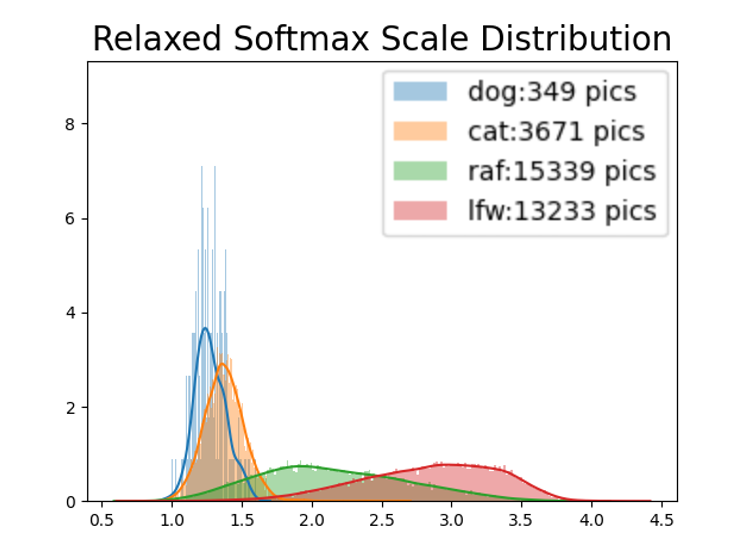}
  \includegraphics[width=0.16\linewidth]{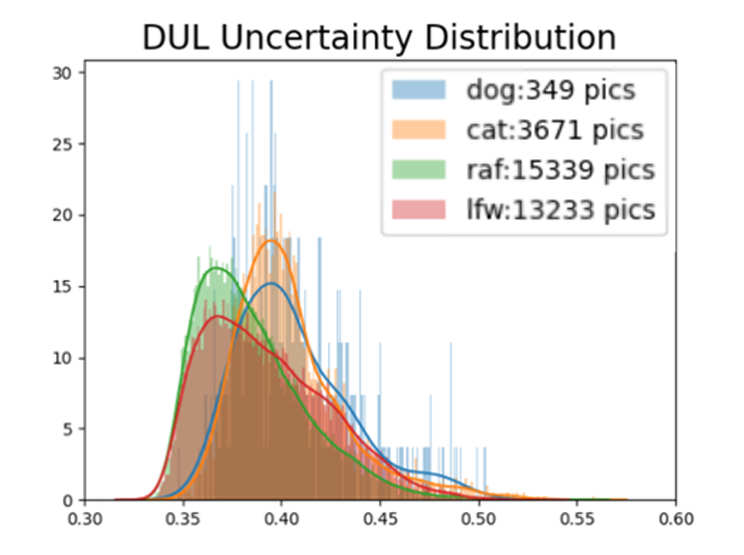}
  \includegraphics[width=0.16\linewidth]{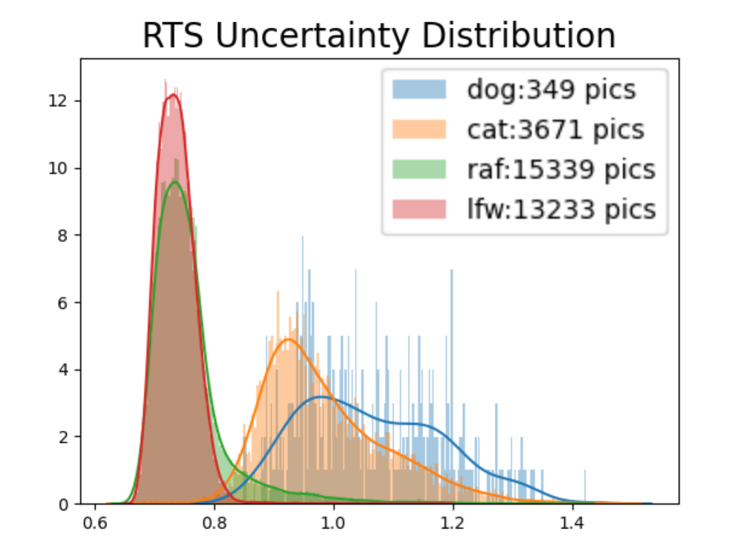}
   \caption{Distributions of magnitudes or uncertainty scores on the OOD dataset. From left to right, each subplot corresponds to the model trained with MagFace, AdaFace, GODIN, Relaxed Softmax, DUL and RTS respectively. LFW and RAFDB are considered as in-distribution data. Cat and dog are considered as out-of-distribution data. All the models are trained on DeepGlint. }
\label{fig:ood_dist}
\end{figure*}

\begin{figure}[t]
\centering
  \includegraphics[width=0.9\linewidth]{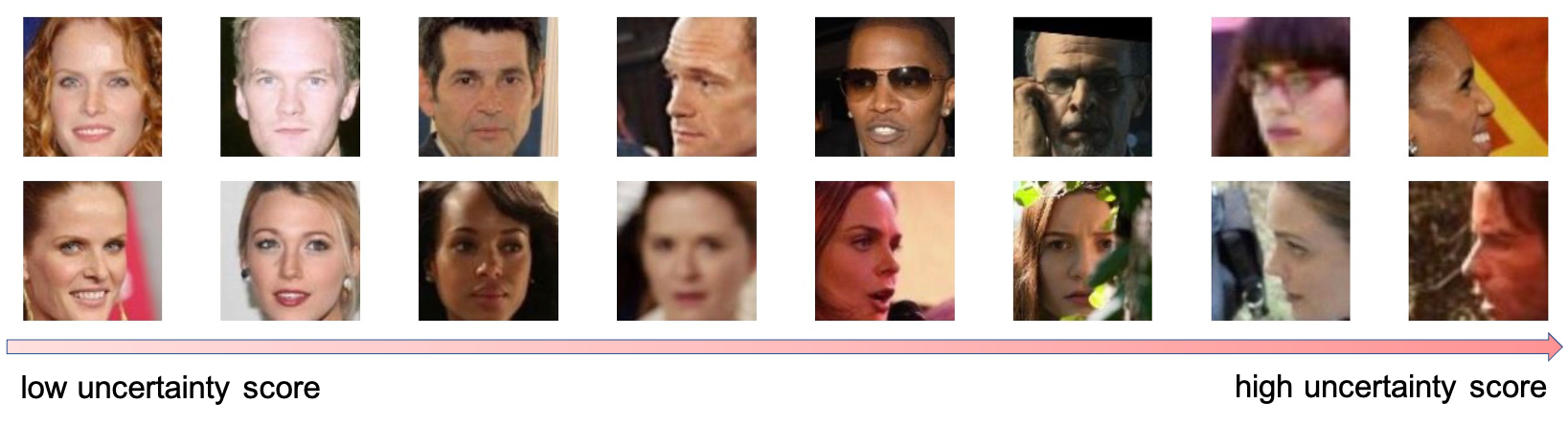}
   \caption{Images with different uncertainty scores estimated by RTS. }
\label{fig:image-score}
\end{figure}

\begin{table}
\centering
\resizebox{0.48\textwidth}{!}{
\begin{tabular}{|l|c|c|c|c|c|}
\hline
pct. & method & LFW & AgeDB & CPLFW & VGG2-FP \\
\hline\hline
\multirow{3}*{0\%} & baseline & 99.38 & 93.95 & 89.88 & 93.88 \\
~ & DUL & 99.33 & 94.02 & 89.18 & 93.80 \\
~ & RTS & \textbf{99.43} & \textbf{94.43} & \textbf{90.55} & \textbf{94.56} \\
\hline
\multirow{3}*{20\%} & baseline & 99.40 & 94.25 & 89.40 & 93.80 \\
~ & DUL & 99.23 & 92.82 & 88.88 & 93.36 \\
~ & RTS & \textbf{99.45} & \textbf{94.42} & \textbf{90.13} & \textbf{94.09} \\
\hline
\end{tabular}
}
\caption{Results of recognition with image noise. Training data: CASIA-WebFace}
\label{tab:image_noise}
\end{table}

\subsection{Training Phase}
\label{ssec:training phase}

\paragraph{Variation of Uncertainty Score.}
To eliminate the effect of dynamic margin for the variation of uncertainty score during the training process of RTS, we use fixed margin in this section to illustrate that our method can adjust the learning strength of clean and noisy samples for stability and accuracy.
As can be seen in Fig.~\ref{fig:variation_uncertainty_curves}, in the early stage, all images have similar high uncertainty scores and the uncertainty score tends to increase as the prediction is inaccurate and random. As the training proceeds, the prediction for high quality data becomes more accurate and the uncertainty score decreases for the data with high confidence while increases for the data with low confidence. In the end, RTS has the ability to distinguish different quality images, giving high-quality images (frontal faces) with low uncertainty and giving low-quality images (profile faces and non-human faces) with high uncertainty.

\paragraph{Training with Noisy Datasets.}
Training on noisy datasets can significantly reduce the accuracy of deep learning models. Learning to fit bad examples indiscriminately will influence the quality of embedding for normal samples.
DUL claims that it can attenuate the effect from ambiguous representations caused by poor quality samples during training. Our learned scale $v(\bm{x})$ can also balance the contributions of different samples in the learning process. The larger $v(\bm{x})$ is, the less the uncertain sample contributes to the gradient of computation. 
We conduct experiments training with noisy CASIA-WebFace to illustrate the robustness of our method. Specifically, we random select different proportions of images from CASIA-WebFace and add Gaussian blur to them. As shown in Table~\ref{tab:image_noise}, RTS is robust to noisy training data.

\begin{table}
\centering
\begin{tabular}{|l|c|c|c|c|c|c|}
\hline
\multirow{2}*{Method} & TNR & TNR & \multirow{2}*{AUC} \\ ~ & @TPR90 & @TPR95 & ~ \\
\hline\hline
MagFace & 79.68 & 63.90 & 97.48 \\
AdaFace & 87.33 & 84.04 & 94.19 \\
GODIN & 59.03 & 27.97 & 89.04 \\
Relaxed Softmax & 96.80 & 76.15 & 96.34 \\
DUL & 39.46 & 31.05 & 66.85 \\
\hline
RTS & \textbf{99.60} & \textbf{98.13} & \textbf{98.38} \\
\hline
\end{tabular}
\caption{Performance of different methods on OOD testset.}
\label{tab:tnr_and_auc}
\end{table}

\begin{figure}
    \centering
    \subfigure[DUL]{\includegraphics[width=0.4\linewidth]{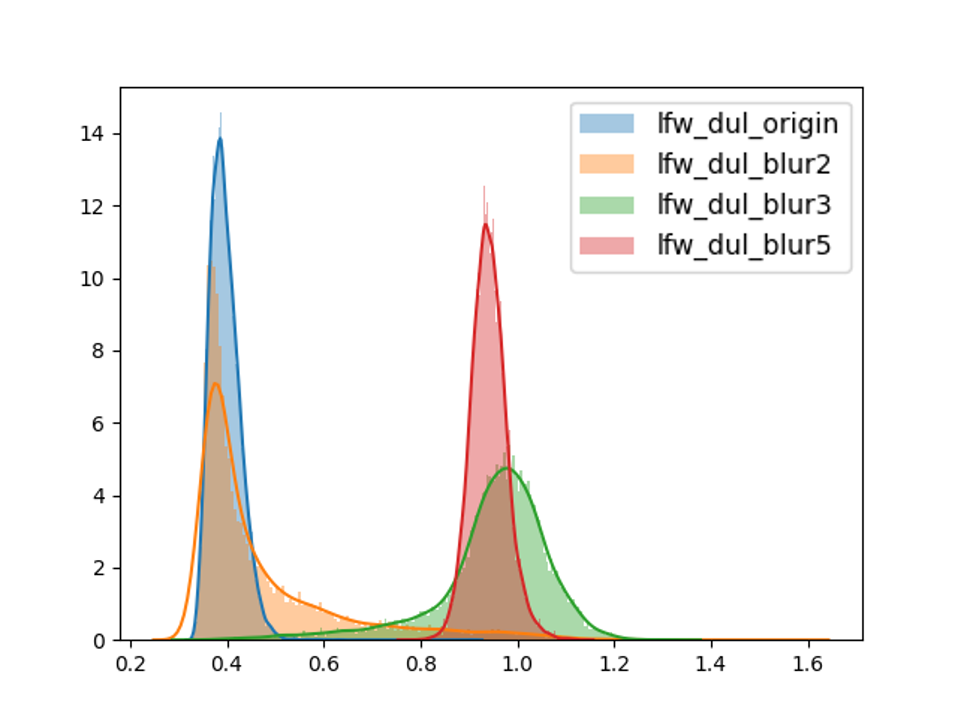}}
    \subfigure[RTS]{\includegraphics[width=0.4\linewidth]{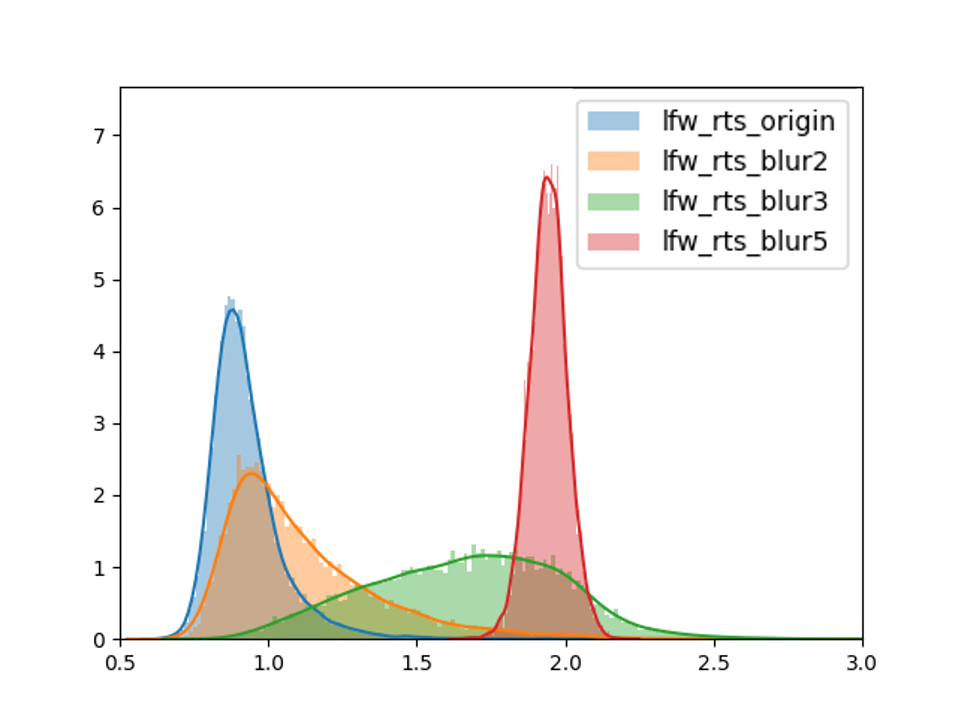}}
    \caption{Uncertainty scores vs. blur levels (Gaussian blur radius are 2, 3 and 5, respectively). See Appendix for distributions of other types of noise. Training data: DeepGlint. }
    \label{fig:noise_levels}
\end{figure}


\begin{table*}
\centering
\begin{tabular}{|l|c|c|c|c|c|c|c|c|c|}
\hline
& Method & LFW & CFP-FF & CFP-FP & AgeDB & CALFW & CPLFW & VGG2-FP & Avg. \\
\hline\hline
\multirow{3}*{w/o uncertainty} & Baseline & 99.80 & 99.67 & 97.95 & 97.90 & 96.07 & 92.58 & 95.90 & \textbf{97.12} \\
~ & MagFace & 99.78 & 99.71 & 97.96 & 97.70 & 95.95 & 92.13 & 95.60 & 96.98 \\
~ & AdaFace & 99.81 & \textbf{99.82} & 97.87 & 97.98 & 96.07 & \textbf{92.83} & 94.96 & 97.05 \\
\hline
\multirow{5}*{w/ uncertainty} & PFE (original) & \textbf{99.82} & - & 93.34 & - & - & - & - & -\\
~ & GODIN & 99.80 & 99.70 & 98.08 & \textbf{98.15} & 95.98 & 91.85 & 95.64 & 97.03 \\
~ & Relaxed Softmax & 99.68 & 99.71 & 97.83 & 97.97 & 95.88 & 92.32 & 95.50 & 96.98 \\
~ & DUL & 99.78 & 99.72 & 97.92 & 97.95 & \textbf{96.15} & 92.66 & 95.22 & 97.06 \\
\cline{2-10}
~ & RTS & 99.77 & 99.74 & \textbf{98.09} & 97.98 & 95.90 & 92.32 & \textbf{96.02} & \textbf{97.12} \\
\hline
\end{tabular}
\caption{Results of recognition trained on DeepGlint (except PFE). The results are all comparably high enough.}
\label{tab:recognition_deepglint}
\end{table*}

\begin{figure}[t]
    \centering
    \includegraphics[width=0.8\linewidth]{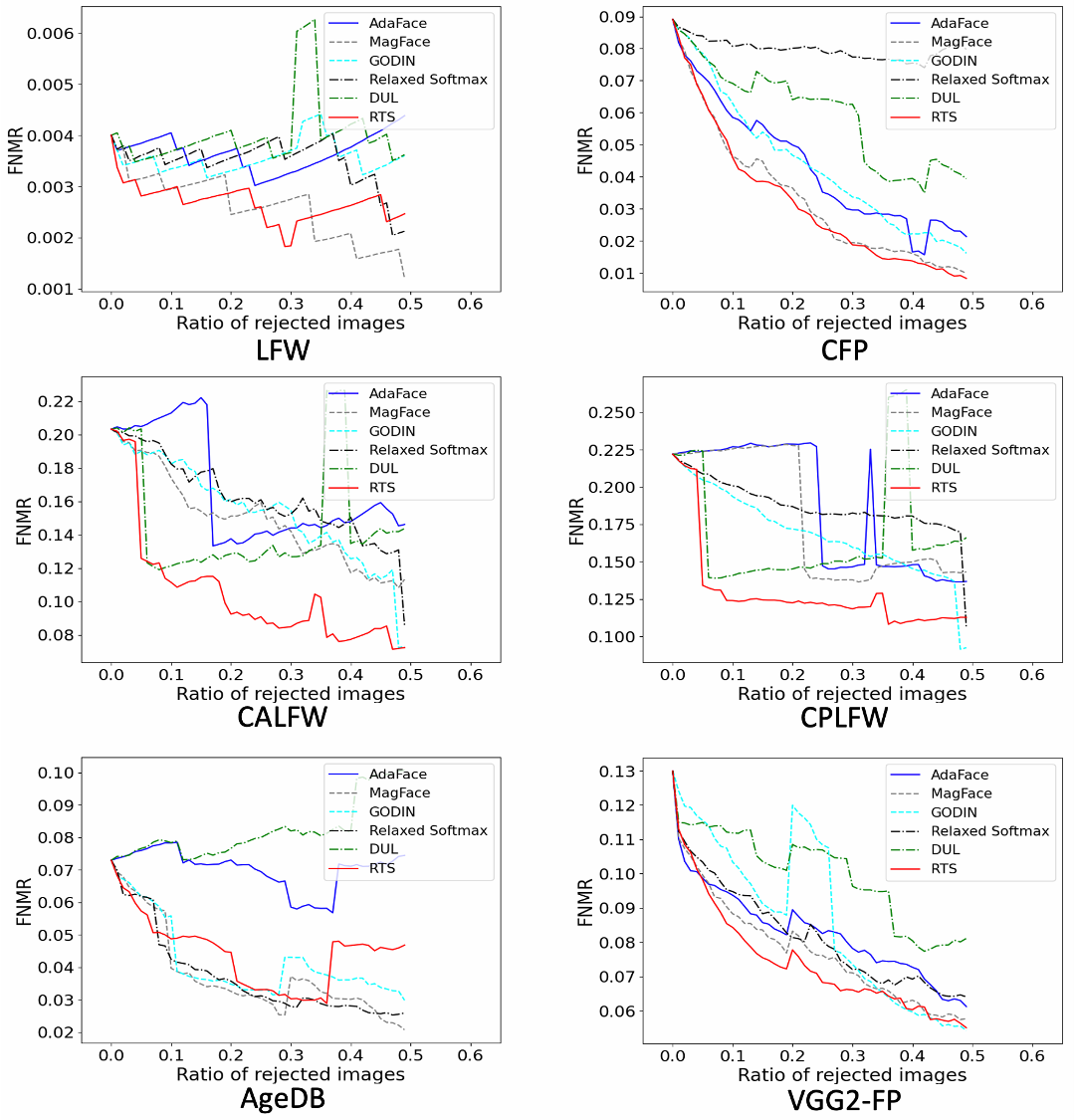}
    \caption{Face recognition performance with rejecting low-quality images. The curves show the effectiveness of rejecting low-quality face images in terms of false non-match rate (FNMR) at false match rate (FMR) threshold of 0.001. Training data: DeepGlint.}
    \label{fig:rej_fr_curve}
\end{figure}

\begin{figure}[t]
\centering
  \includegraphics[width=0.72\linewidth]{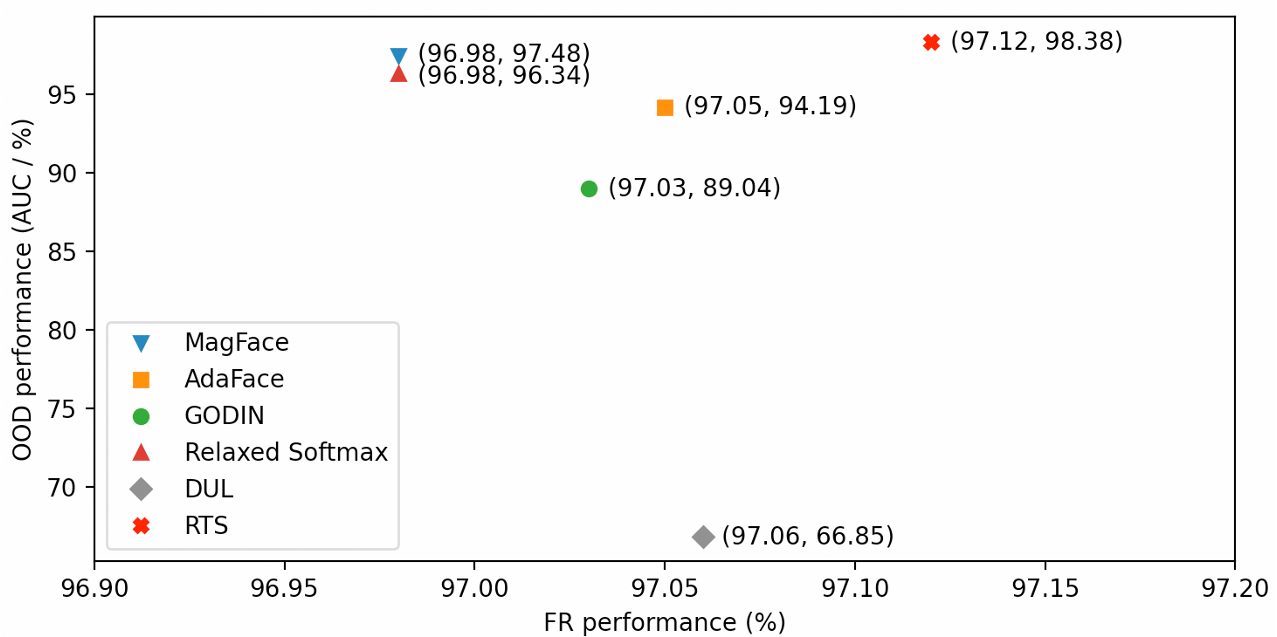}
  \caption{Balance between FR and OOD performance. Training data: DeepGlint.}
\label{fig:FR-OOD}
\end{figure}

\subsection{Testing Phase}
\label{ssec:testing phase}
\paragraph{Relation Between Uncertainty and Image Quality.}
Given the estimated uncertainty scores $v(\bm{x})$ by RTS, we visualize the images with different levels of uncertainty in Fig.~\ref{fig:image-score}. For images with low uncertainty scores, the faces are generally clean and frontal. As the uncertainty level increases, the faces gradually become blurred, occluded, and corrupted by bad lighting conditions and large pose angles. Besides the data uncertainty, the score also reflects the learning difficulty for the input image.

\paragraph{Relation Between Uncertainty and Image Noise.}
To further explore the meaning of the learned uncertainty and its relation to image noise, we evaluate RTS and DUL on noisy test set. As shown in Fig.~\ref{fig:noise_levels}, we corrupt images from LFW with different Gaussian blur radius to obtain new data set LFW-blur. 
The distributions of uncertainty scores are shown in Fig.~\ref{fig:noise_levels} (a - b). As the noise level increases, the predicted uncertainty scores of our RTS gradually shift to large values, while the scores of DUL change irregularly. 
See Appendix for distributions of other types of noise. 
This indicates that the scale of RTS is closely related to the intensity of noise and the quality of image.

\paragraph{Face Recognition with Rejecting Low-quality Images.}
Fig.~\ref{fig:rej_fr_curve} shows the error-versus-reject curves of rejecting different quality face images in terms of false-non-match rate (FNMR). In order to control variables to illustrate the performance of each model on rejection task, we first obtain image features which are used for calculating FNMR from RTS. For non-uncertainty model (MagFace, AdaFace), we use magnitudes of image features obtained from MagFace and AdaFace to reject low-quality images. And for uncertainty models, we use the proposed OOD score in each model (Details can be seen in \ref{ssec:ood}) to reject poor quality samples. Dropping low-quality faces can benefit face recognition performance significantly. As shown in Fig.\ref{fig:rej_fr_curve}, RTS achieves the best FNMR on different ratio of rejected images in all benchmarks except LFW and AgeDB. RTS performs best in the former 20\% of LFW-benchmark and former 8\% of AgeDB-benchmark (only rejecting a small amount of samples), and has high performance uniformity for various testsets. It is noteworthy that in real application scenarios of face recognition, the proportion of low-quality images is usually small and only a small number of samples will be rejected for recognition. Considering this, our method can achieve best performance in real application. RTS is able to distinguish samples with low-quality and give the corresponding sample a reasonable uncertainty score. Compared with other methods, RTS can stably complete rejection task to improve the performance of face recognition.

\subsection{Out-of-Distribution Detection}
\label{ssec:ood}
\paragraph{Uncertainty Scores and Evaluation Metrics.}
To evaluate the performance of uncertainty models on OOD detection task reasonably, we use the proposed score in each model. For GODIN and Relaxed Softmax, the proposed score is the multiplier or denominator to the logit. For DUL, the proposed score is the harmonic mean of estimated variance. We have shown that the estimated value of $v(\bm{x})$ in RTS is related to the scale of temperature, and thus reflects the scale of uncertainty in the classification model. We choose $v(\bm{x})$ as the uncertainty score of our method to detect OOD samples. For non-uncertainty model (MagFace, AdaFace), we use magnitudes of image features obtained from MagFace and AdaFace to detect OOD samples. 
Following literature~\cite{hsu2020generalized}, we use true negative rate at $\{90\%, 95\%\}$ true positive rate (TNR@TPR90, TNR@TPR95) and the area under the receiver operating characteristic curve (AUC) to evaluate OOD detection performance.

\paragraph{OOD Detection Performance.}
Table~\ref{tab:tnr_and_auc} shows that RTS outperforms the other methods on OOD detection task. MagFace, AdaFace and Relaxed Softmax perform well. GODIN has a relatively good OOD result. While DUL cannot detect out-of-distribution data accurately. Besides, the distributions of different models are shown in Fig.~\ref{fig:ood_dist}. We can see that our proposed RTS can better discriminate the in-distribution and out-of-distribution data. Both quantitative and distributions results indicate that RTS is an effective technique to model uncertainty and complete OOD detection task.

\begin{table}
\centering
\begin{tabular}{|l|c|c|c|c|c|c|}
\hline
$\lambda$ & LFW & CFP-FP & AgeDB & CPLFW & VGG2-FP \\
\hline\hline
0.1 & 98.43 & 93.53 & 87.63 & 84.12 & 91.62 \\
1 & 99.23 & 95.57 & 91.97 & 87.03 & 93.28 \\
10 & \textbf{99.43} & \textbf{97.50} & \textbf{94.43} & \textbf{90.55} & \textbf{94.56} \\
\hline
\end{tabular}
\caption{Results of our models trained with different weights for KL divergence ($\delta$ is 16). Training data: CASIA-WebFace.}
\label{tab:kl_weight}
\end{table}

\begin{table}
\begin{center}
\begin{tabular}{|l|c|c|c|c|c|c|}
\hline
$\delta$ & LFW & CFP-FP & AgeDB & CPLFW & VGG2-FP \\
\hline\hline
8 & 99.30 & 97.11 & 93.75 & 89.33 & 94.38 \\
16 & \textbf{99.43} & \textbf{97.50} & \textbf{94.43} & \textbf{90.55} & \textbf{94.56} \\
32 & 99.33 & 97.04 & 93.47 & 89.62 & 93.70 \\
\hline
\end{tabular}
\end{center}
\caption{Results of our models trained with different values for degree of freedom ($\lambda$ is 10). Training data: CASIA-WebFace.}
\label{tab:degree_of_freedom}
\end{table}

\subsection{Face Recognition Performance}
\label{ssec:face recognition}
\paragraph{Face verification accuracy on benchmarks.} We compare RTS with the state-of-the-art uncertainty methods including PFE, GODIN, DUL and Relaxed Softmax, and related non-uncertainty methods including Baseline (ArcFace), MagFace and AdaFace. 
The results of recognition is shown in Table~\ref{tab:recognition_deepglint}. We can see that the recognition performance of RTS is comparable with the state-of-the-art methods on all test sets. This indicates that, besides the ability to reveal uncertainty of images and detect out-of-distribution data, the model trained with RTS can achieve competitive performances in face recognition task. 

\paragraph{Balance Between FR and OOD Performance. } Fig.~\ref{fig:FR-OOD} shows the performance between face recognition (average verification accuracy of all benchmarks) and out-of-distribution detection (AUC) of non-uncertainty and uncertainty models. The comprehensive performances of MagFace, AdaFace and Relaxed Softmax are both close to that of RTS. While MagFace has a great many hyper parameters needed to be adjusted manually and is difficult to reproduce good enough results. Besides, the convergence of Relaxed Softmax is very sensitive to the margin in its prediction head. In comparison, RTS has less hyper parameters and is easier to converge. Our method achieves the best performance in OOD detection task and comparably high enough face verification accuracy, demonstrating that RTS is a unified framework for uncertainty estimation and face recognition.

\subsection{Ablation Study}
\label{ssec:ablation}

\paragraph{Effects of KL divergence.} We study the effects of KL divergence with different weights in this part.
The KL divergence loss works as a regularization term to prevent the uncertainty scale from growing infinitely. When the weight $\lambda < 0.1$, the model have difficulty in converging, and the performance also deteriorates at last.
For large $\lambda (>10)$, the model tends to predict nearly constant variance $v(\bm{x})$, which has little effects in modeling data uncertainty.
We conduct experiments on models with $\lambda \in \{0.1, 1, 10\}$. The results are shown in Table~\ref{tab:kl_weight}.
Through experiments, we find that the model achieves the best performance when $\lambda = 10$. Thus, we set $\lambda = 10$ for our RTS model.

\paragraph{Effects of degree of freedom.} We study the effects of degree of freedom $\delta$, which is a hyperparameter of RTS. Intuitively, $\delta$ determines the shape of density of random temperature, $t$. The results are shown in Table~\ref{tab:degree_of_freedom}. From experimental results, we can see that RTS achieves the best performance when $\delta = 16$. Thus, we set $\delta = 16$ for our RTS model.

\section{Conclusion}
\label{sec:conclusion}

In this paper, we first analysis the connection between temperature scaling and uncertainty modeling in the classification model. Taking a probabilistic view, the temperature scalar is exactly the scale of uncertainty noise implicitly added in the softmax function. Based on this observation, a unified framework, Random Temperature Scaling (RTS), is proposed for uncertainty estimation and face recognition by modeling the uncertainty level by a stochastic distribution. 

Experiments show that RTS can adjust the learning strength of different quality samples for stability and accuracy during training. The magnitude of variance in RTS acts as a metric to reveal the image quality and can be used to detect uncertain, low-quality and even OOD samples in testing phase. 
Face recognition models trained with RTS have higher security and reliability by rejecting untrusted images, especially when deployed in real-world face recognition systems. RTS achieves top performance on both FR and OOD detection tasks. Moreover, models trained with RTS performs robustly on datasets with noise. The proposed module is light-weight and only adds negligible computation cost to the original face recognition model.

\section*{Appendix}
\appendix
\section{Relation between Uncertainty Noise and Softmax with Temperature Scaling}
We show the detailed derivation of the formulas used in softmax and temperature scaling.
Fig.~\ref{fig:temp-scaling} shows a typical classification process. A classification model predicts the score, $s_k$, for each possible class. To make a decision, one can choose the class with the maximum score, which results in a ``Hard'' max method (Fig.~\ref{fig:temp-scaling} top).
However, ``Hard'' max is a deterministic method and cannot reflect the confidence of results. Intuitively, the prediction made from scores (2.1, 1.0, 0.6) should be more confident than that from (2.1, 2.0, 1.3), though the two cases result in the same decision (where class 1 is chosen). 

To take the confidence into account, one can introduce an uncertainty noise, $u_k$, and make a prediction from the random scores $s_k + u_k, k = 1, \dots, C$ (Fig.~\ref{fig:temp-scaling} middle). We assume that $u_k$'s are i.i.d. random variables and follow the standard Gumbel distribution. The probability density function is defined by
\begin{equation}
  p(u_k) = \exp(u_k - \exp(u_k)),
  \label{eq:gumbel_pdf}
\end{equation}
and the cumulative distribution function is defined by
\begin{equation}
  \Pr(u_k \leq x) = \exp(-\exp(-x)).
  \label{eq:gumbel_cdf}
\end{equation}

Given the i.i.d. condition, the probability to predict class $k$ is given by:
\begin{equation}
  \begin{aligned}
    & \Pr(s_k+u_k \geq s_j+u_j, \forall j) \\
    =\ & \int_{-\infty}^{\infty} p(u_k) \cdot \prod_{j\neq k} \Pr(u_j \leq s_k + u_k - s_j | u_k) d u_k \\
    =\ & \int_{-\infty}^{\infty} e^{u_k - \exp(u_k)}\cdot \prod_{j\neq k} e^{-\exp(s_j - s_k - u_k)} d u_k \\
    =\ & \int_{-\infty}^{\infty} e^{- \sum_{j} \exp(s_j - s_k - u_k)}\cdot e^{u_k} d u_k \\
    =\ & \int_{-\infty}^{\infty} e^{- e^{-u_k}\cdot \sum_{j} \exp(s_j - s_k)}\cdot e^{u_k} d u_k.
  \end{aligned}
  \label{eq:prob-class}
\end{equation}
Substituting $z = -e^{-u_k}$ into Eqn.~\ref{eq:prob-class}, we can obtain

\begin{equation}
  \begin{aligned}
    & \Pr(s_k+u_k \geq s_j+u_j, \forall j) \\
    =\ & \int_{-\infty}^{0} e^{z\cdot \sum_{j} \exp(s_j - s_k)} dz \\
    =\ & \left.\frac{1}{\sum_{j} \exp(s_j - s_k)} \cdot e^{z\cdot \sum_{j} \exp(s_j - s_k)} \right|_{-\infty}^{0} \\
    =\ & \frac{1}{\sum_{j} \exp(s_j - s_k)} = \frac{e^{s_k}}{\sum_{j} e^{s_j}}.
  \end{aligned}
  \label{a:eq:prob-softmax}
\end{equation}
Eqn.~\ref{a:eq:prob-softmax} is exactly the softmax function.

We assume that the scale of uncertainty noise can vary by a factor $t$, \ie, $u_k\sim Gumbel(0, t)$. In this case, the new variable $u_k/t$ follows the standard Gumbel distribution. Through the change of variables, we can obtain
\begin{equation}
  \begin{aligned}
    & \Pr(s_k+u_k \geq s_j+u_j, \forall j) \\
    =\ & \Pr(\frac{s_k+u_k}{t} \geq \frac{s_j+u_j}{t}, \forall j) \\
    =\ & \frac{1}{\sum_{j = 1}^{C} \exp(\frac{s_j - s_k}{t})}.
  \end{aligned}
  \label{a:eq:prob-softmax-ts}
\end{equation}
Eqn.~\ref{a:eq:prob-softmax-ts} is exactly the softmax function with temperature scaling, and \textbf{the temperature corresponds to the scale of uncertainty noise added to the classification score} (Fig.~\ref{fig:temp-scaling} bottom). The variance of the uncertainty noise is $\pi^2 t^2 / 6$. The larger the temperature is, the larger the variance of uncertainty becomes. For a small temperature, the noise is close to zero, and the prediction becomes confident.
In knowledge distillation, $t$ is a global value tuned for each task and each model. In confidence calibration, $t(x)$ depends on the input data.

\section{Choice of Temperature Distribution}
\label{sec:temp}

The plain softmax function is a special case of RTS with a constant temperature of 1. In that case, the model is encouraged to learn every sample equally well. 
For RTS, without introducing a bias, we aim to formulate a temperature distribution whose density is concentrated at 1. 
Thus, the proposed Gamma distribution is suitable for the requirement. The temperature is modeled as a sum of independent Gamma variables. The mode of temperature, $v(\bm{x})$, is a learned function of the input image. 
As shown in Fig. 3 of the main paper, $\delta$ determines the peakedness/kurtosis of the distribution, while $v(\bm{x})$ determines the scale of the distribution. $v(\bm{x})$ is further constrained by the regularization term (Eqn. 6 of the main paper) to be close to 1. 
During training, RTS can adjust the learning strength of clean and noisy samples to achieve high stability and accuracy. The noisy and OOD samples can be gradually filtered out via the uncertainty score (Fig. 4 of the main paper).

\section{More Experimental Results}
\label{sec:exp}

\begin{figure}
\centering
  \includegraphics[width=0.39\linewidth]{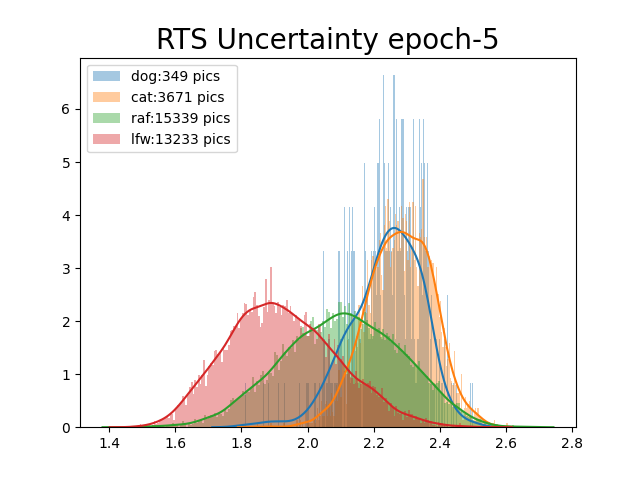}
  \includegraphics[width=0.39\linewidth]{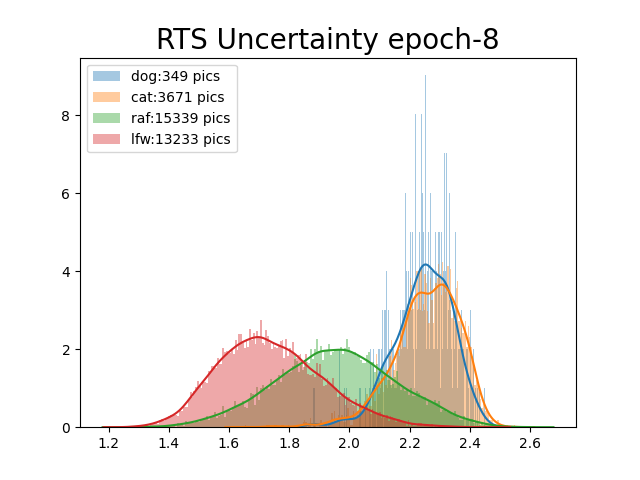}\\
  \includegraphics[width=0.39\linewidth]{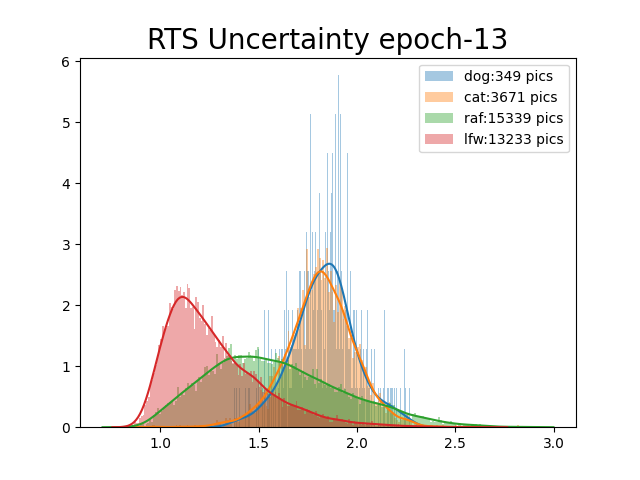}
  \includegraphics[width=0.39\linewidth]{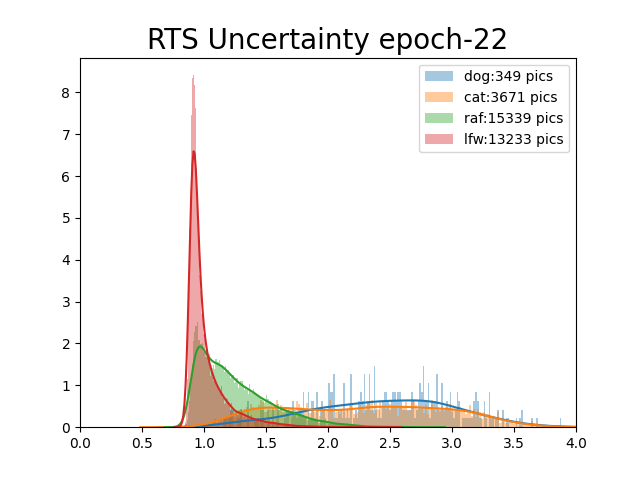}
  \caption{The dynamic distributions of learned variance $v(\bm{x})$ in RTS with fixed margin trained on DeepGlint~\cite{deepglint} at epoch 5, 8, 13, and 22.}
\label{fig:dynamic-dist}
\end{figure}

\paragraph{Variation of Uncertainty Score.}
Variation of uncertainty score during training phase can be seen in Fig.~\ref{fig:dynamic-dist}. In the early stage, the uncertainty score tends to increase as the prediction is inaccurate and random. As the training proceeds, the prediction for high quality data becomes more accurate and the uncertainty score decreases for the data with high confidence while increases for the data with low confidence. 

\begin{figure}
\centering
  \subfigure[DUL distributions of LFW-pepper (density: 0.03 / 0.05 / 0.1)]{\includegraphics[width=0.39\linewidth]{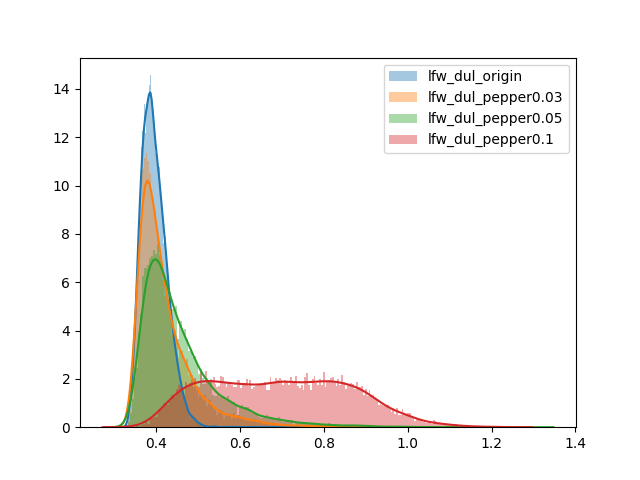}}
  \hspace{10mm}
  \subfigure[RTS distributions of LFW-pepper (density: 0.03 / 0.05 / 0.1)]{\includegraphics[width=0.39\linewidth]{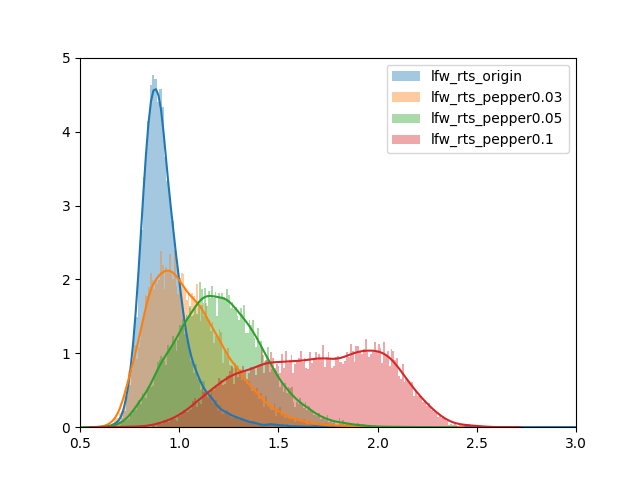}}
  \subfigure[DUL distributions of LFW-rotate (rotate degree: 15 / 30 / 60)]{\includegraphics[width=0.39\linewidth]{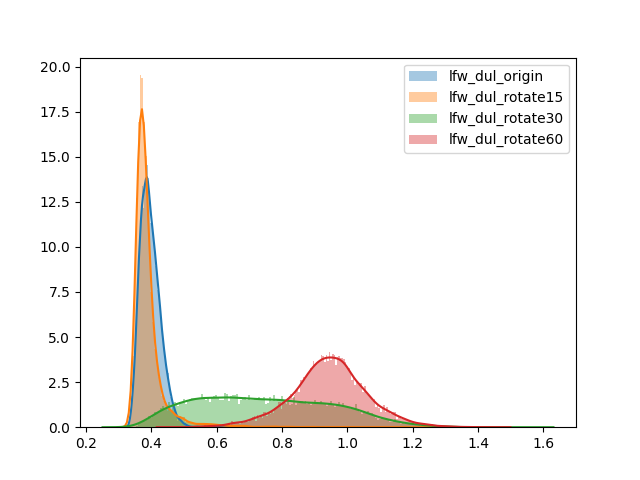}}
  \hspace{10mm}
  \subfigure[RTS distributions of LFW-rotate (rotate degree: 15 / 30 / 60)]{\includegraphics[width=0.39\linewidth]{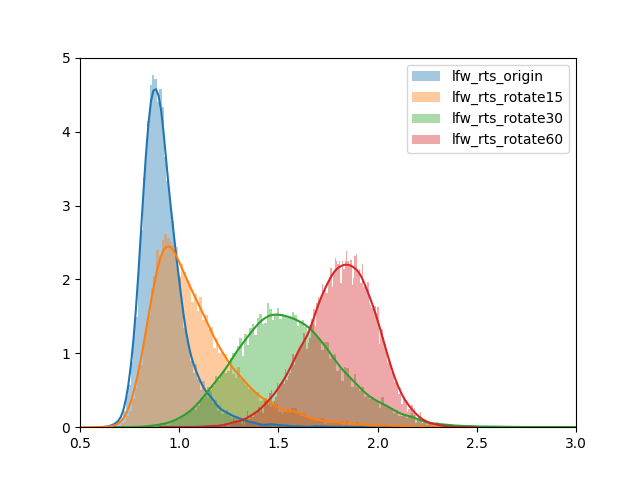}}
  \subfigure[DUL distributions of  LFW-mask (side length: 20 / 40 / 60)]{\includegraphics[width=0.39\linewidth]{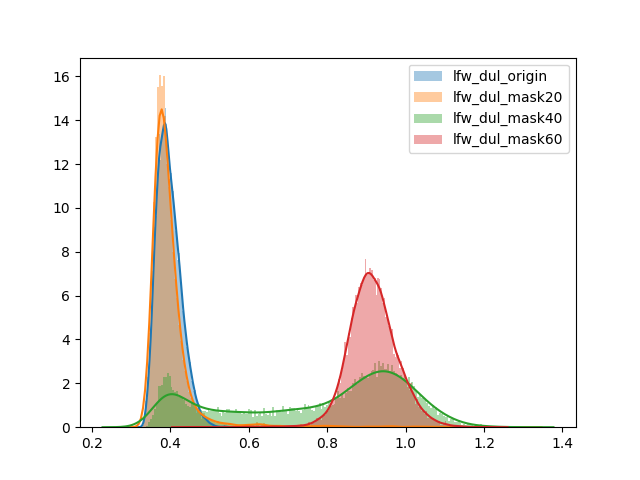}}
  \hspace{10mm}
  \subfigure[RTS distributions of  LFW-mask (side length: 20 / 40 / 60)]{\includegraphics[width=0.39\linewidth]{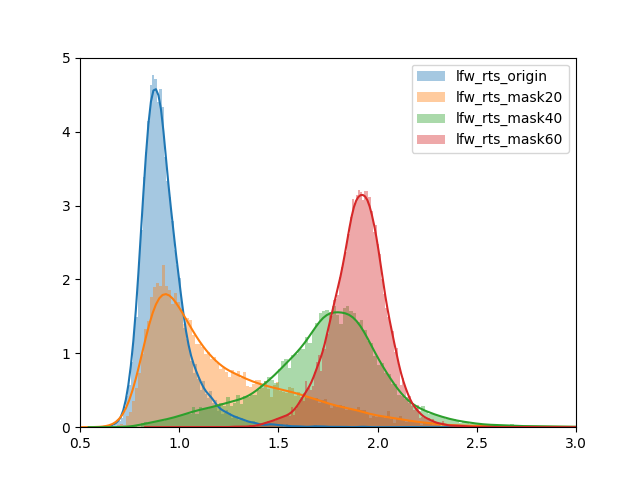}}
  \subfigure[A sample of LFW and corresponding noisy images.]{\includegraphics[width=0.6\linewidth]{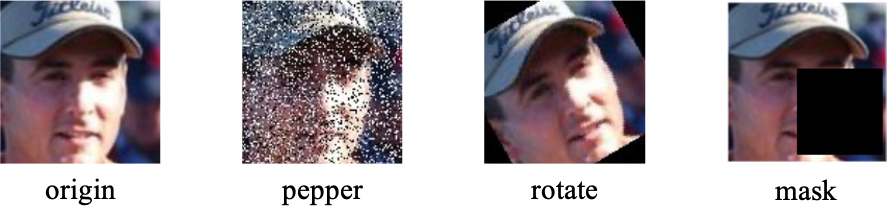}}
  \caption{(a - f): Uncertainty scores vs. noise levels. The left and right columns correspond to the distributions of DUL and RTS. Training data: DeepGlint. (g): A sample of original and noisy images.}
\label{a:fig:noise_levels}
\end{figure}

\paragraph{Relation Between Uncertainty and Image Noise.}
To further explore the meaning of the learned uncertainty and its relation to image noise, we evaluate RTS and DUL~\cite{chang2020data} on more noisy test set. As shown in Fig.~\ref{a:fig:noise_levels} (g), we corrupt images from LFW~\cite{huang2008labeled} using different method (pepper, rotate, mask) to obtain new data set LFW-noise (LFW-pepper, LFW-rotate, LFW-mask). 
The distributions of uncertainty scores for different types of noise are shown in Fig.~\ref{a:fig:noise_levels} (a - f). As the noise level increases, the predicted uncertainty scores of our RTS gradually shift to large values, while the scores of DUL change irregularly.

\section*{Acknowledgments}
This work was supported in part by the National Natural Science Foundation of China under Grant (62272450,U1813218), the Joint Lab of CAS-HK,  in part by  the Shanghai Committee of Science and Technology, China (Grant No. 20DZ1100800).
%
This work was supported by Alibaba Group through Alibaba Innovative Research Program.

\bibliography{aaai23}

\end{document}